\documentclass[10pt,twocolumn,reqno]{article}

\usepackage{cvpr}
\usepackage{times}
\usepackage{epsfig}
\usepackage{graphicx}
\usepackage{amsmath}
\usepackage{amssymb}

\usepackage{subfiles}

\graphicspath{
 {./images/}
}

\usepackage{rotating}
\usepackage{float}
\usepackage{adjustbox}
\usepackage{enumitem}

\setlist[1]{itemsep=-4pt, leftmargin=10pt}\usepackage{array}
\newcolumntype{P}[1]{>{\centering\arraybackslash}p{#1}}
\usepackage[pagebackref=true,breaklinks=true,letterpaper=true,colorlinks,bookmarks=false]{hyperref}

\def\cvprPaperID{****} 

\ifcvprfinal\fi

\cvprfinalcopy 

\def\cvprPaperID{****} 

\ifcvprfinal\fi
\title{Understanding the Decision Boundary of Deep Neural Networks: \\ An Empirical Study}

\author{David Mickisch, Felix Assion, Florens Gre\ss ner, Wiebke G\"unther, Mariele Motta \\ 
neurocat GmbH \\
{\tt\small dmi@neurocat.ai, fa@neurocat.ai}
}
\begin{document}

\maketitle

\begin{abstract}
Despite achieving remarkable performance on many image classification tasks, state-of-the-art machine learning (ML) classifiers remain vulnerable to small input perturbations. Especially, the existence of adversarial examples raises concerns about the deployment of ML models in safety- and security-critical environments, like autonomous driving and disease detection. Over the last few years, numerous defense methods have been published with the goal of improving adversarial as well as corruption robustness. However, the proposed measures succeeded only to a very limited extent. This limited progress is partly due to the lack of understanding of the decision boundary and decision regions of deep neural networks. Therefore, we study the minimum distance of data points to the decision boundary and how this margin evolves over the training of a deep neural network. By conducting experiments on \textsc{Mnist}, \textsc{Fashion-Mnist}, and \textsc{Cifar-10}, we observe that the decision boundary moves closer to natural images over training. This phenomenon even remains intact in the late epochs of training, where the classifier already obtains low training and test error rates. On the other hand, adversarial training appears to have the potential to prevent this undesired convergence of the decision boundary. 
\end{abstract}


\section{Introduction}

Deep learning has facilitated technological advances in a variety of domains, e.g.\ computer vision, natural language processing, and robotics. However, the notable success in achieving high test performances should not obscure the fact that we are still lacking a sufficient understanding of the functioning of deep neural networks. \newline 
In the past years, it has become apparent that deep neural networks are quite brittle, in particular they are vulnerable to small - even imperceptible - perturbations of the input \cite{Akhtar, Dodge}. This observation at least partially contradicts the common belief that deep networks generalize well to unseen, similar examples. These harmful inputs can be the result of distributional shifts or general noise in the environment of the classifier \cite{Volpi}, e.g.\ unusual lighting or weather conditions for an autonomous car. Alternatively, they might be adversarial examples, i.e.\ perturbed natural images that were intentionally crafted by some adversary to cause misclassifications. In the past, these two types of harmful perturbations have been analyzed by mostly separate research communities, namely the corruption robustness and the adversarial robustness researchers. Recently it has become more and more evident that the vulnerability to these different perturbations is closely connected and should therefore not be analyzed separately \cite{Hendrycks}. For example, in \cite{Ford} the authors derived a promising estimate for the size of small worst-case adversarial perturbations with the help of the test error on additive Gaussian noise.  \newline 

With the seminal work of Biggio et al.\ \cite{Biggio4} and Szegedy et al.\ \cite{Szegedy} as a starting point, there has been tremendous research effort directed towards exploring methods that generate adversarial examples \cite{Assion, Eykholt} as well as defenses that aim at increasing robustness \cite{Guo, Metzen2}. The consistent success of strong targeted adversarial attacks in computer vision, like PGD \cite{Madry} and C\&W \cite{Carlini2}, suggests that an adversary can basically create any desired classification output by adding a suitable adversarial perturbation to the natural image. This has also been impressively shown by Metzen et al.\ \cite{Metzen} in the context of semantic segmentation. Here, the authors were able to create universal, i.e.\ input-agnostic, adversarial perturbations that lead to any desired target segmentation. These adversarial attack results bring us to the realization that the decision boundary of a conventional deep neural network seems to be close to almost every input image. In other words, the closeness of the decision boundary to natural data points explains the vulnerability of state-of-the-art ML models to certain input perturbations.  \newline
A vast number of adversarial defenses therefore try to increase the minimal distance, also called margin in the input space, of natural images to the decision boundary. Among other things, researchers have proposed regularization penalties \cite{Gu, Cisse}, data augmentation techniques \cite{Tramer, Papernot3} and specific architectures \cite{Papernot4, Schott} to obtain a more desirable course of the decision boundary, and in consequence to get to more robust classifiers. Unfortunately, the majority of defenses have not succeeded and were "broken" shortly after their publication due to stronger adversaries \cite{Athalye2}. 
In general, PGD adversarial training is still viewed as the most reliable and the most successful adversarial defense \cite{Madry}. But, it should be noted that adversarial training shows promising results under very limited threat models and tends to overfit on specific attacks instead of improving general robustness \cite{Kang}. \newline

This limited progress in increasing robustness has led more and more researchers to the question whether robustness can be obtained at all by deep neural networks \cite{Fawzi, Shafahi, Schmidt}. The well-known, but rather theoretical, statement that neural networks with a single hidden layer are universal function approximators has partly given us a false sense of security. For example, recent work indicates that commonly used neural network topologies with a large number of relatively low-dimensional hidden layers may not lead to universal function approximators \cite{Johnson}. 
Additionally, publications have shown empirically, as well as theoretically, that the decision regions of modern ML classifiers, i.e.\ the regions of the input space that lead to a certain output class, tend to be connected sets \cite{Johnson, Fawzi2}. 
This strong topological restriction on the decision regions might already limit the expressive power of these models and thus, their maximal achievable robustness. \newline
In addition, there have been efforts to derive general robustness bounds for classes of classification problems, independent of the used classification function. Fawzi et al.\ \cite{Fawzi} provide fundamental upper bounds on the achievable robustness assuming that the natural data comes from a smooth generative model. Under this assumption on the origin of the data, they show that any type of classifier is prone to adversarial perturbations as long as the latent space of the generative model is high-dimensional. \newline

Overall, we are still at an early stage of understanding the decision making and limitations of deep neural networks. Especially, the course of the decision boundary and factors that influence its course have to be explored in more detail. Findings in this area can then guide us towards promising adversarial defenses as well as general robustness bounds. In this paper, we want to continue along this path by providing an empirical study focusing on the distance of data points to the decision boundary and how this margin evolves over the training of the classifier. To the best of our knowledge, the change of the margin during the training process has not yet been analyzed in the existing literature. Our experiments with neural network classifiers on \textsc{Mnist}, \textsc{Fashion-Mnist}, and \textsc{Cifar-10} lead us to three central observations: 

\begin{itemize}
\item The decision boundary moves closer to training as well as test images over training. This convergence of the decision boundary even continues in the late phases of training where the neural network already obtains low training and test error rates.
\item Adversarial training results in a significantly different development of the decision boundary. Here, the average distance to the decision boundary of the images stays at a relatively high level over training. The clear downward trend observable for standard training is damped considerably, which underlines the success of adversarial training in improving robustness for simple classification tasks.
\item Wrongly classified images from the natural data distribution are on average significantly closer to the decision boundary than correctly classified data points. During training the decision boundary is pushed towards these points, which implies that their already small distance to the decision boundary decreases even further with increasing epoch number. This observation holds for adversarial as well as standard training.
\end{itemize}
  
Due to the success of adversarial attacks on ML models, it is not surprising that the decision boundary is close to the majority of natural data points after training. But, our findings still challenge common beliefs about the training of neural network classifiers. It does not seem to be true that training moves the decision boundary away from the training data in order to facilitate generalization, or at least this is not true for all directions in the input space. 

\section{Background} \label{sec: related_work}

In this section, we formalize the notion of decision boundary, and summarize related work on the decision boundary of deep neural networks. For our experiments it will be crucial to calculate the minimal distance of an image to the decision boundary of the classifier. Unfortunately, calculating the exact margin in the input space is in general intractable, thus one has to make use of a reasonable upper bound. We will use DeepFool \cite{Dezfooli} for this margin approximation. Due to its significant role within the following empirical study, we will recall the central intuition underlying the DeepFool adversarial attack in this section.

\subsection{The Decision Boundary}

We define a classifier as a function $f: \mathbb{R}^n \rightarrow \mathbb{R}^c$, where $n$ denotes the number of dimensions in the input space (e.g.\ number of pixels in an image) and $c$ is the number of classification classes. For an input point $x \in \mathbb{R}^n$ the output $f(x)$ can be interpreted as the vector of softmax values of the classifier. The classification decision is then given by 
$$\hat{k}(x) = \underset{k=1,...,c} {\operatorname{argmax}} \; f_k(x).$$
With this notation, we can now define the decision boundary $\mathcal{D} \subseteq \mathbb{R}^n$ of $f$ as the set
\begin{align*}
\begin{split}
\mathcal{D} := \{ x \in \mathbb{R}^n  \mid \exists & k_1, k_2 = 1,...,c, \; k_1 \neq k_2, \\
& f_{k_1}(x) = f_{k_2}(x) = \operatorname{max}_k f_k(x)\}.
\end{split}
\end{align*}

In other words, these are the points where the decision of the classifier is tied. 
The margin $d_2(x) \in \mathbb{R}_{\geq 0}$ of a data point $x$ is then given by
\begin{align} \label{eq: 1}
\begin{split}
d_2(x) = \underset{\delta \in \mathbb{R}^n}{\min} &\; \| \delta \|_2 \\
&s.t. \; x + \delta \in \mathcal{D}
\end{split}
\end{align}

In the above margin definition, we make use of the $\ell_2$-norm, but one can define the margin with respect to any reasonable distance measure. In our empirical study, we will also consider the margin with respect to the $\ell_{\infty}$-norm, which we will denote $d_{\infty}$.   \newline
The existence of adversarial examples for state-of-the-art neural networks indicates that natural images lie near the decision boundary $\mathcal{D}$ with high probability, hence $d_2(x)$ is small for most data points $x$. At the same time, deep neural networks also tend to be rather robust to random noise \cite{Yu}. It may thus be concluded that the decision boundary is close in a "few" directions, but further away with respect to the majority of perturbation directions. 
In \cite{He2} the authors confirm this observation by adding randomly sampled orthogonal directions to benign as well as adversarial images. For the benign images, these sampled perturbations rarely change the classification decision. On the other hand, they observe that most of the adversarial examples generated by the FGSM attack \cite{Good1} are not at all robust to these random distortions.  \newline
For our experiments, we do not want to rely on sampled perturbations for the calculation of $d_2(x)$, but rather utilize an optimization method. Since the margin optimization problem (\ref{eq: 1}) is intractable for deep neural networks, one has to search for an approximate solution. To be more precise, one searches for a small perturbation $\delta$ such that $\hat{k}(x+\delta) \neq \hat{k}(x)$. Ideally, we then have $d_2(x) \approx \| \delta \|_2$, although $x+ \delta$ is not necessarily an element of $\mathcal{D}$. \newline
We will obtain the margin estimate with the help of DeepFool, but one can also make use of other strong adversarial attacks for the generation of $\delta$. For example, in \cite{Nar} the authors use the PGD attack to approximate the distance of training data to the decision boundary. They find that the usual cross-entropy loss is one contributing factor to small margins and that a differential training procedure leads to more robust models. \newline
However, DeepFool has proven to generate particularly small adversarial perturbations which makes it suitable for margin approximation. Jiang et al.\ \cite{Jiang} used a simplified targeted version of DeepFool with a single iteration step to estimate $d_2(x)$ for images of the training data set. These distances then formed the basis for a measure which correlates with the generalization gap of a trained neural network. \newline

Apart from determining the distance to the decision boundary, it is also desirable to understand more general geometric properties of the decision boundary. Already in the early adversarial robustness literature, it has been hypothesized that the decision boundary of a deep neural network locally resembles the decision boundary of a linear classifier. In \cite{Good1} the authors claim that this "too" linear behavior explains the success of the FGSM adversarial attack which utilizes the linearization of the loss function of the network.  More recently, Fawzi et al.\ \cite{Fawzi2} empirically showed that the decision boundary near natural images is flat in most directions, and curved only in very few directions. Furthermore, their results suggest that the decision boundary is biased towards negative curvatures. \newline

\subsection{DeepFool}

DeepFool is an untargeted, iterative adversarial attack which stops as soon as a perturbation $\delta$ has been found with $\hat{k}(x_0+\delta) \neq \hat{k}(x_0)$ for some given data point $x_0$. Moosavi-Dezfooli et al.\ \cite{Dezfooli} introduced DeepFool with the goal to provide a method that can calculate adversarial perturbations with similar efficiency as FGSM (or comparable attacks like BIM \cite{Kurakin} and PGD \cite{Madry}), but which at the same time leads to a more accurate approximation of the robustness of $f$ at $x_0$. The authors achieve this by making use of the well-understood orthogonal projection mapping of a point onto the decision boundary of an affine classifier. \newline

To be more specific, in every iteration step of DeepFool the class probability functions $f_k, \; k = 1,...,c,$ of the classifier $f$ are linearized around the current position $x_i$. Then, one calculates the smallest perturbation $\delta_i$ with respect to the $\ell_2$-norm which moves $x_i$ onto the decision boundary of the linearized model of $f$. This perturbation $\delta_i$ can be written down in closed form:
$$ \delta_i := \frac{ |f_{\hat{l}}(x_i) - f_{\hat{k}(x_0)}(x_i) |}{\| \nabla f_{\hat{l}}(x_i) - \nabla f_{\hat{k}(x_0)}(x_i) \|_2^2 } (\nabla f_{\hat{l}}(x_i) - \nabla f_{\hat{k}(x_0)}(x_i))$$
with index
$$\hat{l}=\hat{l}(x_i):= \operatorname{argmin}_{k \neq \hat{k}(x_0) } \frac{|f_{\hat{l}(x_i)}(x_i) - f_{\hat{k}(x_0)}(x_i)|}{\| \nabla f_{\hat{l}(x_i)}(x_i) - \nabla f_{\hat{k}(x_0)}(x_i)  \|_2 }. $$
Now, let $N$ denote the stopping index of this iterative scheme for image $x_0$, i.e.\  $\hat{k}(x_N) \neq \hat{k}(x_0)$. Then, the desired adversarial perturbation is given by
$$ \delta = \sum_{i=0}^{N-1} \delta_i.$$
Overall, the DeepFool attack can be viewed as a function $\operatorname{DeepFool}: \mathbb{R}^n \rightarrow \mathbb{R}^n$ which takes an image $x_0$ as input and returns a corresponding adversarial perturbation $\delta$ for the given classifier $f$.
In the original paper, the authors also formulate adaptations of the DeepFool algorithm to any $\ell_p$-distance measure for $p \in [1, \infty]$. Since we also want to consider the margin with respect to the $\ell_{\infty}$-norm in our empirical study, we will use the $\ell_{\infty}$ adaptation for these approximations. For the closed form formula of $\ell_{\infty}$-DeepFool we refer to the original paper \cite{Dezfooli}. \newline

\section{Experimental Results}

The objective of the experiments is to track the $\ell_2$-norm margin values $d_2$ as well as the $\ell_{\infty}$-norm margin values $d_{\infty}$ for training data as well as test data over the training process of a deep neural network. We obtain these approximate distances of images to the decision boundary with the $\ell_2$-DeepFool algorithm and the $\ell_{\infty}$-DeepFool algorithm, respectively. Hence, we assume that $$d_2(x) = \| \operatorname{DeepFool}_2(x) \|_2$$ and $$d_{\infty}(x) = \| \operatorname{DeepFool}_{\infty}(x) \|_{\infty}$$ for any image $x$.
In order to derive general observations, we will analyze the average margin and the corresponding standard error over large parts of the training as well as the test data set in every training epoch. To be more precise, for an image data set $\mathcal{D}$ we report
$$d_2^{avg}:= \frac{1}{|\mathcal{D}|}\sum_{x_i \in \mathcal{D}} d_2(x_i), \; \; d_{\infty}^{avg}:= \frac{1}{|\mathcal{D}|}\sum_{x_i \in \mathcal{D}} d_{\infty}(x_i)$$
and 
\begin{align*}
\begin{split}
d_2^{se}:= \frac{1}{|\mathcal{D}|}\sqrt{\sum_{x_i \in \mathcal{D}}(d_2(x_i)-d_2^{avg})^2}, \\ 
d_{\infty}^{se}:=\frac{1}{|\mathcal{D}|}\sqrt{\sum_{x_i \in \mathcal{D}} (d_{\infty}(x_i)-d_{\infty}^{avg})^2},
\end{split}
\end{align*}
where $| \mathcal{D}|$ denotes the cardinality of the image set $\mathcal{D}$. For one of the following observations we will also plot the distributions $d_2(x)$ and $d_{\infty}(x)$ for all images $x \in \mathcal{D}$ in several training epochs. These distribution visualizations help us to better understand how the different images contributed to the calculated average margin and standard error. \newline
It should be noted that we only use successful adversarial perturbations for the margin approximation. Thus, if the DeepFool attack is not able to find a small adversarial perturbation for a given image, we iteratively perturb the image with Gaussian noise and retry DeepFool on the perturbed image until we find a successful adversarial example. \newline   

To ensure consistency of our observations, we train several classifiers with different architectures and computer vision tasks. In particular, we conduct experiments on \textsc{Mnist}, \textsc{Fashion-Mnist}, and \textsc{Cifar-10}. In the following figures we will present the experimental results for a convolutional neural network trained on the \textsc{Fashion-Mnist} data set. The given graphs summarize our results for the first $40$ epochs of training. After this limited training period, the \textsc{Fashion-Mnist} model obtains good, but not yet competetive, error rates. However, all our central observations can already be made during the first $40$ epochs, and they continue to hold when we extend the training time.  
Comparable graphs for different architectures and for the other tasks - \textsc{Mnist} and \textsc{Cifar-10} - can be found in Appendix \ref{sec: A}. Overall, our experiments suggest three major findings which we will discuss separately in the following sections. 

\subsection*{Observation 1: Standard Training}

\begin{figure*}[t]
\begin{center}
\centerline{%
      \includegraphics[width=0.35\textwidth]{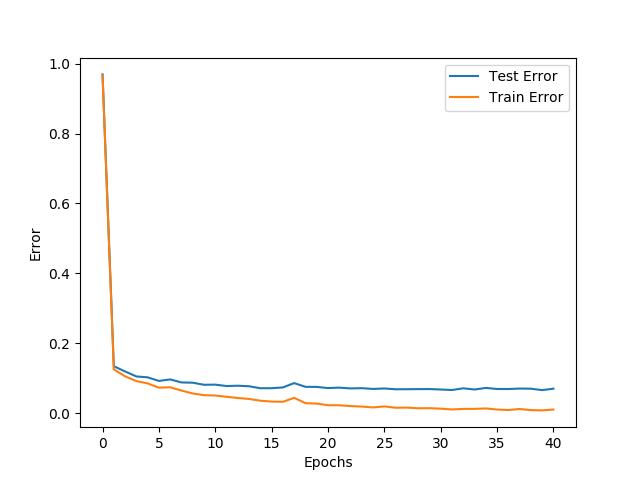}%
    \includegraphics[width=0.35\textwidth]{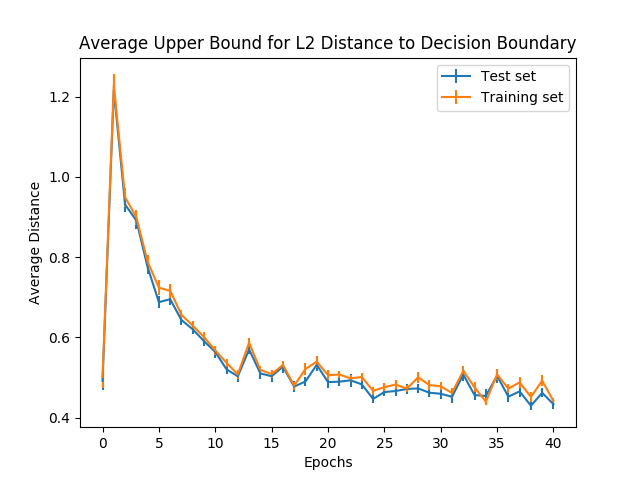}%
      \includegraphics[width=0.35\textwidth]{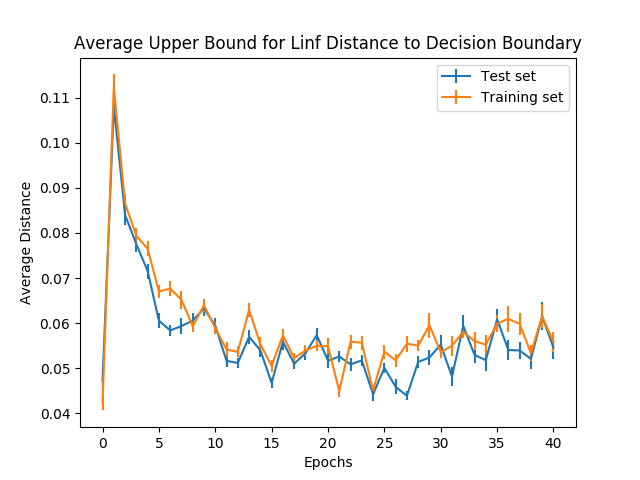}%
      }%
\end{center}
   \caption{Experimental results of a convolutional neural network (CNN) on the \textsc{Fashion-Mnist} dataset: (1) Left: Development of training and test error; (2) Middle: Development of average $\ell_2$-margin $d_2^{avg}$ and standard error $d_2^{se}$ over $1000$ randomly picked images from the training and the test data set; (3) Right: Development of average $\ell_{\infty}$-margin $d_{\infty}^{avg}$ and standard error $d_{\infty}^{se}$ over $1000$ randomly picked images from the training and the test data set.} 
 \label{fig: results_vanilla}
\end{figure*}

Due to the vulnerability of state-of-the-art neural networks to adversarial examples, we know that at least a large portion of the training images as well as test images lie close to the decision boundary after training. As a consequence, we expect low values for the average $\ell_2$-norm margin $d^{avg}_2$ and the average $\ell_{\infty}$-norm margin $d^{avg}_{\infty}$ for a trained classifier. But it is still unclear how the margin metrics $d^{avg}_2$ and  $d^{avg}_{\infty}$ evolve over the training process. To analyze this, we train a convolutional and a dense architecture with cross-entropy loss for the three given computer vision tasks. Sample results for the convolutional neural network (CNN) on \textsc{Fashion-Mnist} are shown in Figure \ref{fig: results_vanilla}.
In general, we observe a similar development of the average margins and their standard errors during the training process for all models and tasks. \newline

After the weight initialization - i.e.\ before the first training weight update - test and training images are very close to the decision boundary. After the first epoch, $d^{avg}_2$ as well as $d^{avg}_{\infty}$ jump to a higher level. Thus, already the first training epoch changes the course of the decision boundary decisively, although it does not yet lead to a classifier with optimal train and test accuracy. 
In the subsequent epochs, the average margins have a clear downward trend and they never get back to the peak of the first epoch. Especially in the early training epochs, the level of $d^{avg}_2$ and $d^{avg}_{\infty}$ drops significantly. The margins decrease less strictly in later phases of the training, but they still decrease noticeably.
Overall, we see a strong negative correlation between the average distance to the decision boundary and the training (or test) accuracy. At the same time, the standard errors $d_2^{se}$ and $d_{\infty}^{se}$ remain relatively small and stable throughout the whole training. It should also be noted that there is no significant difference between the margin values of the training and the test set. Hence, the decision boundary does not appear to be closer to test images compared to training images. \newline

From these results, we can derive the general finding that the decision boundary moves closer to training as well as test images during training. 
This observed convergence of the decision boundary is an undesirable side effect of training, since it shows that trained models with state-of-the-art test accuracy will end up with relatively low average margins. In particular, increasing the training time of a model might lead to a decrease in robustness. \newline

Unfortunately, we can not yet offer a clear and provable explanation for this behavior of the decision boundary during training. At first glance, one might try to justify this observation by assuming a lack of model capacity. In other words, the chosen model architectures might just not be able to simultaneously achieve good accuracies as well as sufficient decision boundary margins. The existence of this trade-off between performance and robustness would then automatically imply a decrease of the margins during training, since the cross-entropy function forces the network to achieve high accuracy in every training epoch. \newline 
Alternatively, one might be tempted to view this phenomenon as a sign of overfitting on the training data, because overfitting also leads to a decision boundary which is getting unnecessarily close to natural data points. But, both hypotheses are hard to validate, and we even notice aspects of our experimental results contradicting these ideas. For example, the distances to the decision boundary decline throughout phases of the training where the test error is also decreasing significantly. This contradicts the overfitting explanation, because overfitting would manifest itself in a downturn of the test accuracy. On the other hand, the following section and its related experiments will show that the exact same model architectures can also lead to a totally different development of the decision boundary, which confutes the lack of model capacity explanation.  \newline
Thus, the question remains why this phenomenon of decreasing average margins occurs and whether this movement of the decision boundary can be prevented.

\subsection*{Observation 2: Adversarial Training}

\begin{figure*}[t]
\begin{center}
\centerline{%
      \includegraphics[width=0.35\textwidth]{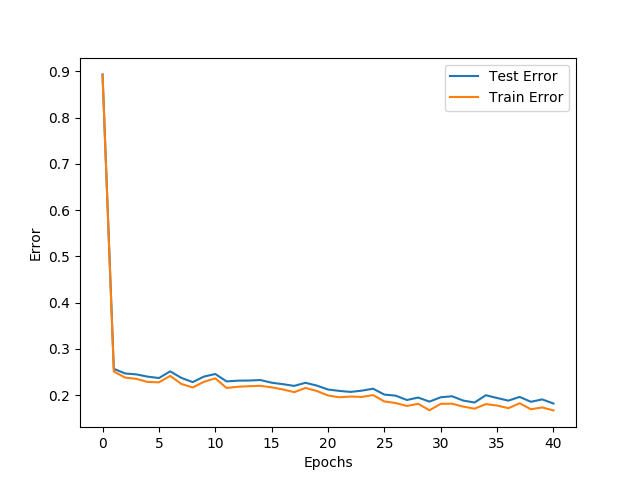}%
    \includegraphics[width=0.35\textwidth]{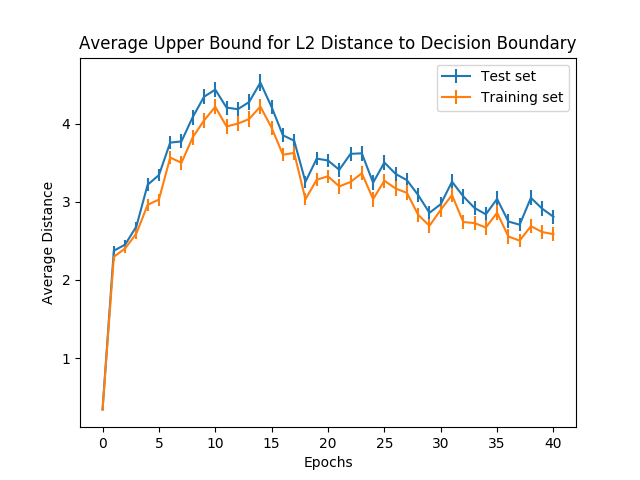}%
      \includegraphics[width=0.35\textwidth]{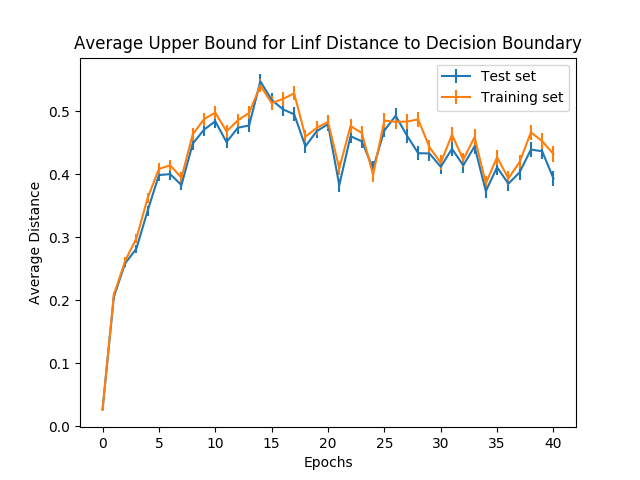}%
      }%
\end{center}
   \caption{Experimental results of a convolutional neural network (CNN) with PGD adversarial training on the \textsc{Fashion-Mnist} dataset: (1) Left: Development of training and test error; (2) Middle: Development of average $\ell_2$-margin $d_2^{avg}$ and standard error $d_2^{se}$ over $1000$ randomly picked images from the training and the test data set; (3) Right: Development of average $\ell_{\infty}$-margin $d_{\infty}^{avg}$ and standard error $d_{\infty}^{se}$ over $1000$ randomly picked images from the training and the test data set.} 
 \label{fig: results_adversarial}
\end{figure*}

We again train convolutional and dense neural networks on \textsc{Mnist} and \textsc{Fashion-Mnist}, but this time perform adversarial training. We keep the cross-entropy loss function and network architectures as in the previous experiments. In every training batch, one now replaces 50\% of the images by adversarial images generated by the PGD attack. Then, we again track the average distances of natural train as well as test images to the decision boundary with respect to the $\ell_2$- and the $\ell_{\infty}$-norm. In Figure \ref{fig: results_adversarial} we summarize the results for the CNN architecture on \textsc{Fashion-Mnist}. \newline
As before, one notices a jump of the average margins after the first epoch of training compared to the level at network initialization. However, in the following epochs, we observe significant deviations from the results of standard training.
After the first epoch, $d^{avg}_2$ as well as $d^{avg}_{\infty}$ remain at a relatively high level compared to standard training. Even though a slight downward trend is observable, it is not as severe as in the setting considered before. It can also be noticed that this downward trend starts in later epochs compared to standard training. Our experimental results for \textsc{Mnist} even show a slight upward trend of the average margins with training time (see: Appendix \ref{sec: A}).
At the same time, the standard errors $d_2^{se}$ and $d_{\infty}^{se}$ are comparable to the past experiments, i.e.\ remain at a rather stable level throughout training for all computer vision tasks. \newline

We come to the conclusion that the previously observed steady decrease of the average margins is not an inevitable phenomenon. The injection of PGD adversarial examples into the training set leads to a decision boundary which is further away from natural data points in comparison to a decision boundary of a classically trained model. Furthermore, the experiments suggest that the robustness of the adversarially trained classifier does not significantly degrade throughout training time. At least for these simple computer vision tasks, we can therefore confirm the general belief that PGD adversarial training creates comparably robust models. \newline 

A further interesting aspect is the high level of average margin size measured in the $\ell_2$-norm, although the PGD attack is concerned with finding small $\ell_{\infty}$-norm adversarial perturbations. Hence, the $\ell_{\infty}$-based adversarial training procedure was also able to increase the minimal $\ell_2$-distance to the decision boundary for a large number of training and test images. It is still an open research question to which extent adversarial training can be used to robustify classifiers against broad classes of adversarial attacks. Past studies suggested that adversarial training does not transfer well between imperceptibility metrics, in particular that $\ell_{\infty}$-based adversarial training does not necessarily lead to robustness with respect to other $\ell_p$-norms, e.g.\ \cite{Kang}. Our results indicate that there actually exist settings, where the positive impact of adversarial training transfers to broader classes of perturbations.

\subsection*{Observation 3: Wrongly Classified Images}

\begin{figure*}[t]
\begin{center}
\centerline{%
   \includegraphics[width=0.35 \textwidth]{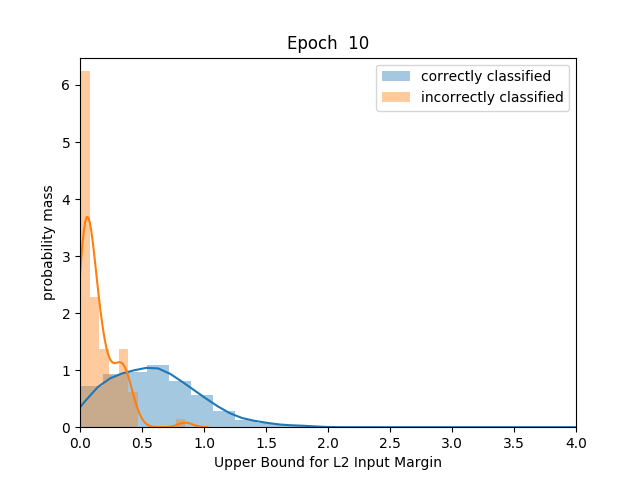}%
    \includegraphics[width=0.35 \textwidth]{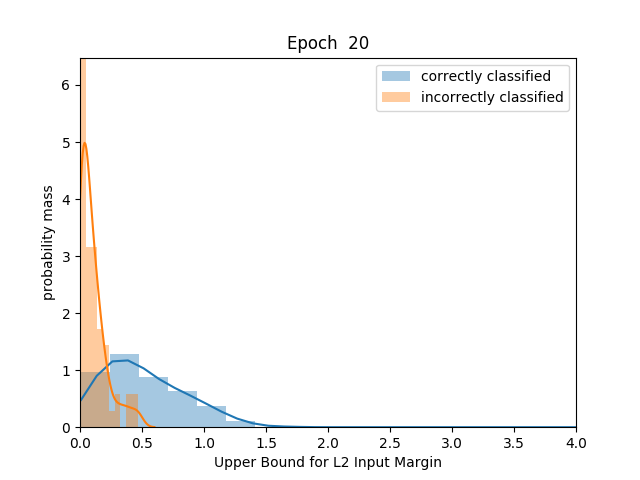}%
    \includegraphics[width=0.35 \textwidth]{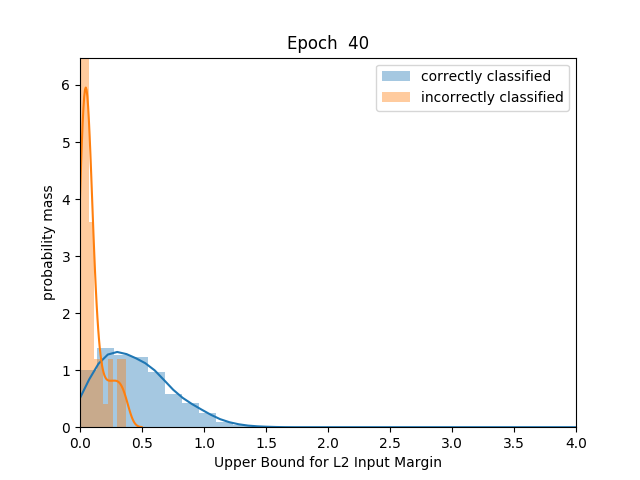}%
    }%
\centerline{%
\includegraphics[width=0.35 \textwidth]{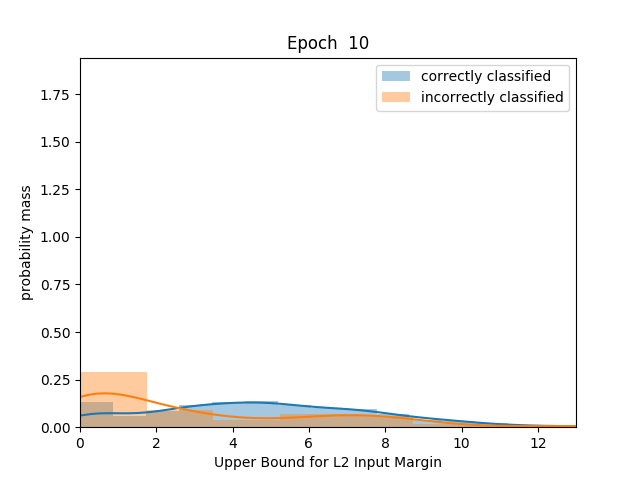}%
\includegraphics[width=0.35 \textwidth]{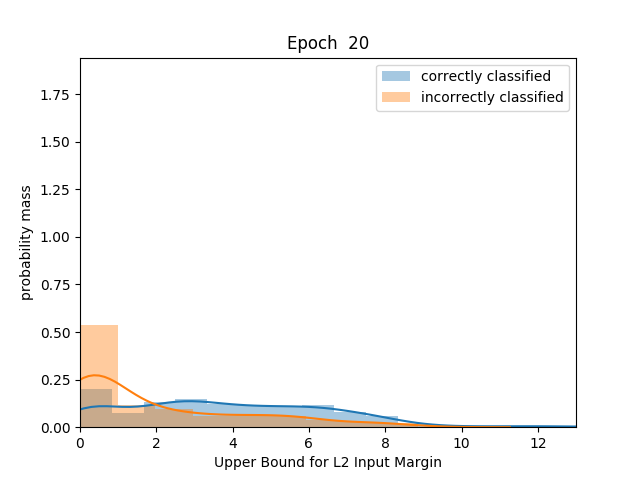}%
\includegraphics[width=0.35 \textwidth]{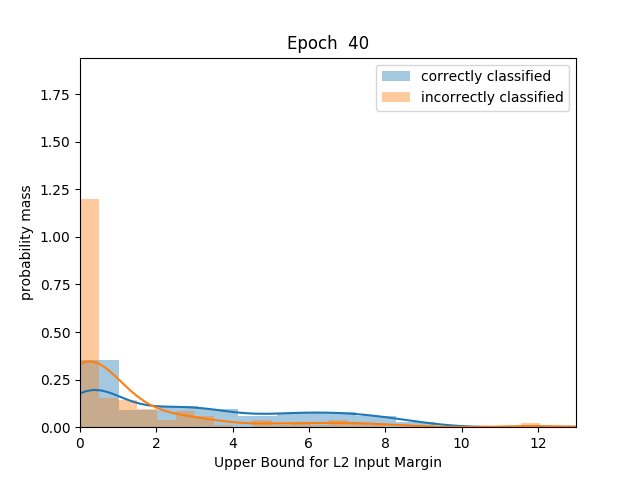}%
}%
\end{center}
   \caption{Sample $L^2$-margin distributions for $1000$ training images and the training epochs $10, 20$ and $40$: (1) First row: $d_2(x)$ distribution for the classically trained CNN on \textsc{Fashion-Mnist}; (2) Second row: $d_2(x)$ distribution for the adversarially trained CNN on \textsc{Fashion-Mnist}.} 
 \label{fig: distributions}
\end{figure*}

\begin{figure*}[t]
\begin{center}
\centerline{%
   \includegraphics[width=0.5 \textwidth]{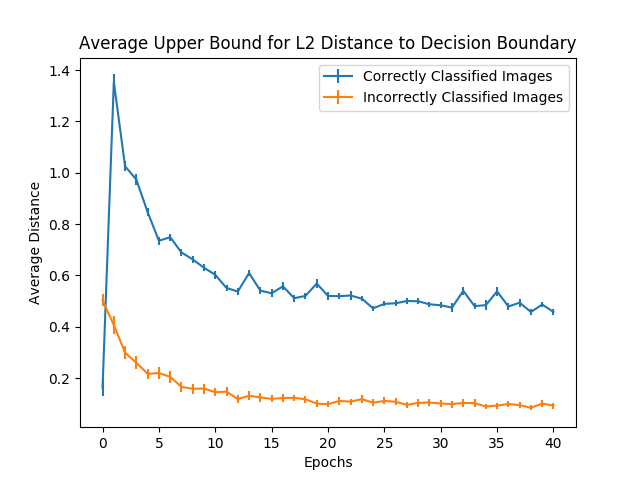}%
    \includegraphics[width=0.5 \textwidth]{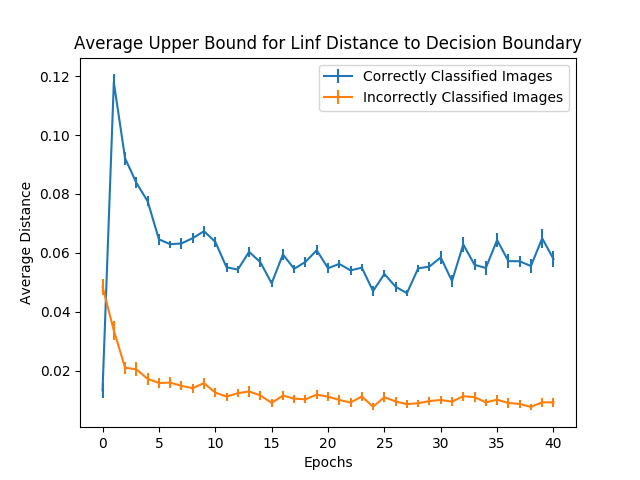}%
    }%
\end{center}
   \caption{Development of average margins and standard errors of the classically trained CNN on \textsc{Fashion-Mnist}. For the calculation of these values we used $1000$ randomly picked images of the test set: (1) Left plot: development of $d_2^{avg}$; (2) Right plot: development of $d_{\infty}^{avg}$.} 
 \label{fig: average_margin_split}
\end{figure*}

Up until now, we have primarily focused on average decision boundary margins and standard errors during the training process. These aggregated statistical values only provide a limited understanding of the distances of training and test images to the decision boundary. We, therefore, analyze the distribution of $d_2(x)$ and $d_{\infty}(x)$ for images $x$ from the training as well as the test set at different training epochs. We again consider the classically trained models and the models trained with PGD adversarial training. Figure \ref{fig: distributions} shows sample distributions of $d_2(x)$ for the classically and the adversarially trained CNN on \textsc{Fashion-Mnist}. \newline 
Additionally, we separate correctly and incorrectly classified images in these distribution graphs, which directly brings us to one apparent realization. The images that were assigned to the wrong class by the deep neural network tend to lie significantly closer to the decision boundary than correctly classified ones. The calculated margins also have a smaller empirical variance (or standard error), hence they are rather concentrated around the mean of the margin approximates. \newline

The distribution of correctly classified images and the distribution of incorrectly classified images wander more and more in the direction of the origin during standard training of a model. This underlines again that the decision boundary moves closer to natural images over the training process (see also: Figure \ref{fig: average_margin_split}). As a consequence, images which were wrongly classified by a fully trained model will be comparably easy to push over the decision boundary via a small perturbation. In the future, this observation might be useful for the detection of misclassified, natural images. \newline

\begin{figure}[t]
\begin{center}
   \includegraphics[width=0.95\linewidth]{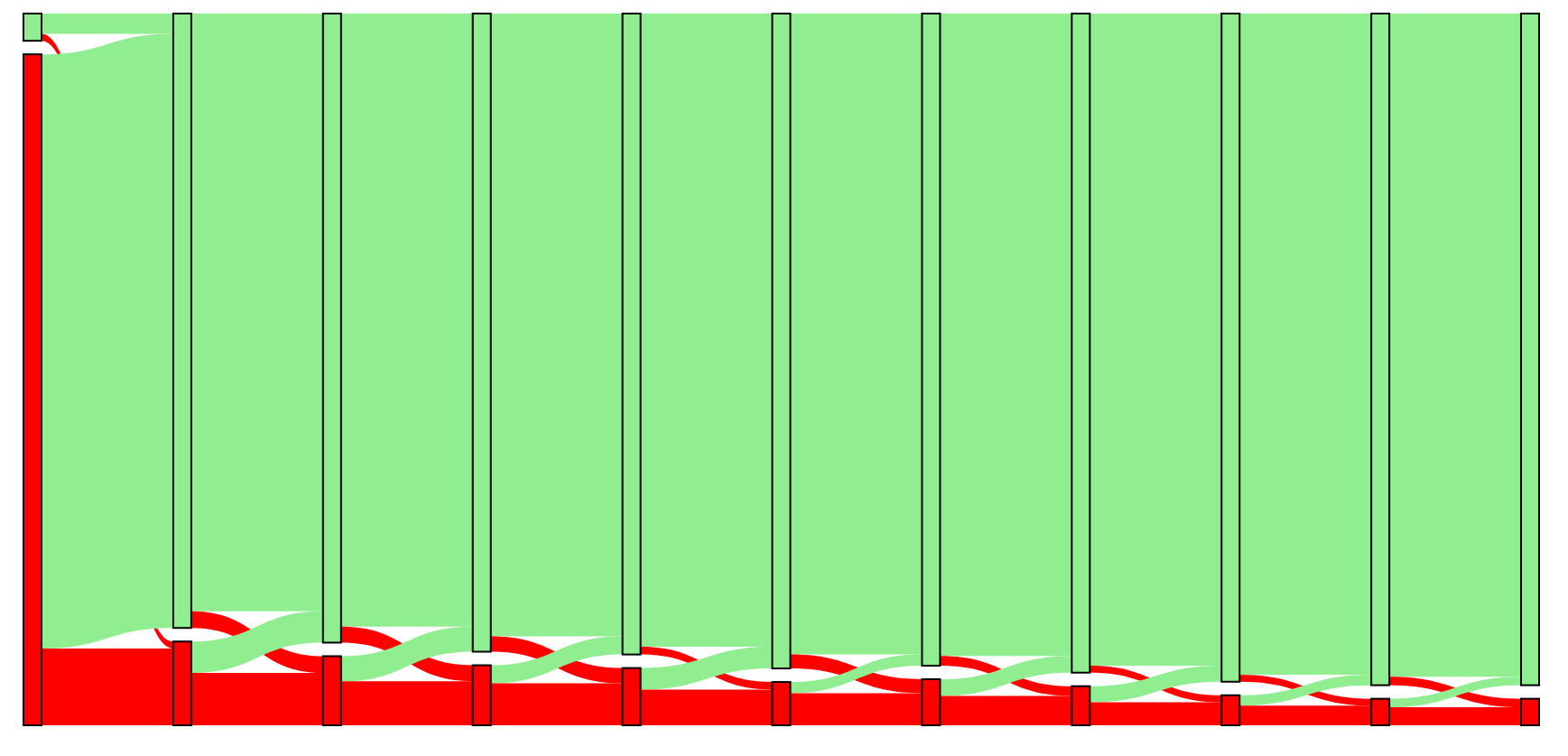}
\end{center}
   \caption{This Sankey plot visualizes the movement of natural images between correct and incorrect classification for the classically trained CNN over the different epochs of training. The plot is based on the whole \textsc{Fashion-Mnist} training set.}
 \label{fig: sankey}
\end{figure}

A possible explanation for this large margin difference between correctly and incorrectly classified images is the fact that the cross-entropy loss punishes wrong decisions during training. Thus, the decision boundary is pushed towards these mistakes, in order to turn these wrong decisions into correctly classified images. This directly implies that the distance to the decision boundary of these wrongly classified images decreases from one epoch to the next, i.e.\ the general movement of the decision boundary towards natural data points is reinforced. This hypothesis is also supported by the given Sankey diagrams in Figure \ref{fig: sankey}. Here, we observe that the majority of incorrectly classified images in one training epoch has also been incorrect in the prior epoch. Hence, the average margin values and the margin distributions of the wrongly classified images of two adjacent epochs rely largely on the same images and thus, allow the statement that the decision boundary is pushed towards these images. \newline
The Sankey plots also show that the decision boundary eventually reaches large parts of the initially misclassified images. These images become correctly classified images, which also results in a smaller training and test error rate. This automatically leads to new correctly classified images with a low distance to the decision boundary in every training epoch. Since the decision boundary has just reached these images and is thus still very close, they have a negative impact on the average margin of the correctly classified images. However, it should be noted that the different distribution graphs also indicate a general trend of the correctly classified images towards smaller margins, which can not all be attributed to the switch of incorrectly classified images to correctly classified images. \newline

\begin{figure}[t]
\begin{center}
   \includegraphics[width=0.95\linewidth]{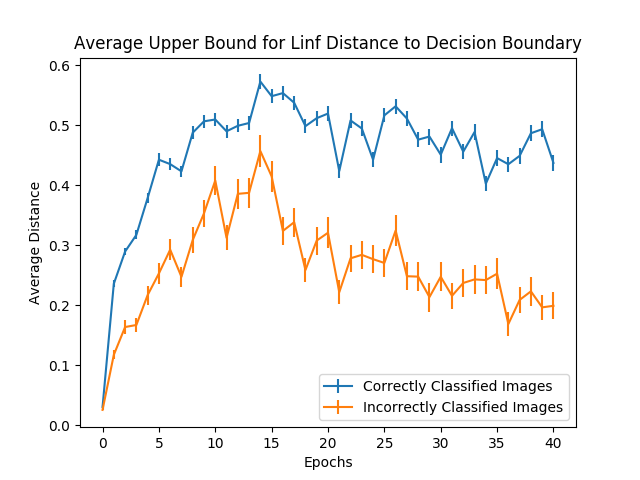}
\end{center}
   \caption{Development of $\ell_{\infty}$ average margin $d_{\infty}^{avg}$ and standard error $d_{\infty}^{se}$ of adversarially trained CNN on \textsc{Fashion-Mnist}. For the calculation of these values we used $1000$ randomly picked images of the test set.}
 \label{fig: average_margin_split_adversarial}
\end{figure}

It is not surprising that adversarially trained models still show the same phenomenon of decreasing margins for incorrectly classified images, while at the same time being able to stabilize the average margins of the training and test set. In Figure \ref{fig: average_margin_split_adversarial} we see that the average $\ell_{\infty}$-margins for the incorrectly classified images have a clear downward trend after the first training epochs. It becomes obvious that the decision boundary of an adversarially trained classifier tries to balance two goals encoded in the loss function and training procedure. It tries to achieve a high accuracy on the training set and simultaneously, to be robust in the neighborhood of training images. The decreasing margins of wrongly classified images resemble the attempt to increase accuracy, while the increasing, or at least stable, margins of correctly classified images show the desired increase in robustness.

\section{Future Work \& Conclusion}
In this empirical study, we have seen that the decision boundary of a state-of-the-art deep neural network moves closer to training and test images during training. The movement of the decision boundary even continues in late phases of training, although the test and training accuracy barely changes at this point. This leads to the conclusion that fully trained models are susceptible to adversarial perturbations as well as general corruption noise. \newline
On the other hand, adversarial training has the potential to prevent this undesired downward trend of the distances to the decision boundary. In general, the average margin of training and test data is at a higher level in this adapted training procedure. Besides, we even noticed a transfer of robustness between the $\ell_2$-norm and the $\ell_{\infty}$-norm for the \textsc{Mnist} and \textsc{Fashion-Mnist} task. \newline
Furthermore, for trained classifiers there exists a significant difference between correctly and incorrectly classified images concerning their distances to the decision boundary. Incorrectly classified images lie a lot closer to the decision boundary than correctly classified images, and here the decision boundary comes closer to these images over training, too. This observation remains present for both standard and adversarial training. \newline

This empirical study contributes to a better understanding of the decision boundary, but there are still a lot of open research questions - even related to the above results. Our observations put the widely studied problem of deep neural networks being susceptible to adversarial examples into a different light. The vulnerability itself indicates that the decision boundary of neural networks is close to most natural images after training. However, we have found that this property of neural networks is not predetermined by their initialization or architecture. Rather, this insufficiency is created during training. Therefore, from our perspective, it seems most promising to further study the influence of loss functions and training procedures on the margins. \newline
Nevertheless, in future experiments one has to analyze whether the given observations hold for more varying network architectures and complex computer vision tasks. It is also crucial to double check the exactness of the DeepFool margin approximates. At this point, one could make use of other strong adversarial attacks, or even apply formal verification techniques to derive lower bounds for the margins of data points. These further investigations will hopefully help us to identify provable explanations for the observed phenomena. \newline

Recently, poor calibration has also been proposed as a potential contributing factor to the robustness issues of ML models. It has been argued that, due to bad uncertainty estimates, the network is not able to identify shifts in the data distribution, that is, out-of-sample instances \cite{bradshaw2017adversarial, smith2018understanding}. This claim is supported by \cite{li2017dropout} that find better calibrated models are able to detect certain classes of adversarial examples. Therefore, it is interesting to also explore the connection between calibration and the development of the distance to the decision boundary over training in more detail. In our third observation we have already seen that the distance to the decision boundary can be a helpful uncertainty metric. There is a significant difference between correctly and incorrectly classified images concerning their margins, although the network usually assigns high confidence scores to both classes of images - which is a clear sign of poor calibration. \newline

In general, we hope that more researchers pick up on the idea to track changes of the decision boundary throughout training, instead of solely concentrating on fully trained networks. This adds a new dimension to the robustness evaluation of a classifier and gives us a better chance to detect different causes of insufficient adversarial and corruption robustness.

{\small
\bibliographystyle{ieee} 
\bibliography{egbib}
}
\clearpage

\onecolumn

\appendix
\section{Results for \textsc{mnist}} \label{sec: A}

\begin{figure*}[h]
\begin{center}
\centerline{%
      \includegraphics[width=0.5\textwidth]{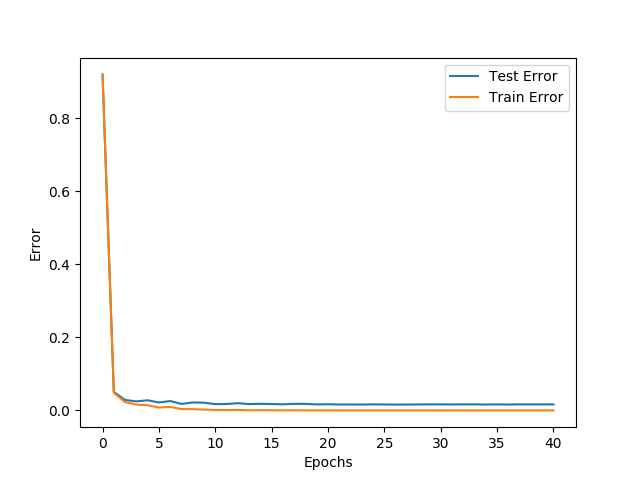}%
    \includegraphics[width=0.5\textwidth]{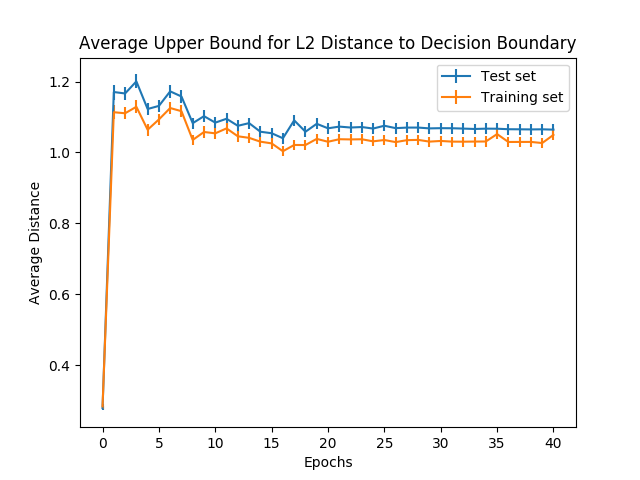}%
      }%
\end{center}
   \caption{In (a) on the left, we observe that the train and test error both converge quickly to almost $0$. \newline
   In (b) we see that even though there is little room for improvement in the test error, the upper bound to the 
   distance to the decision boundary is decreasing in a small but visible 
   fashion after the first epoch supporting observation 1.} 
 \label{fig: mnist_results_vanilla}
\end{figure*}

\begin{figure*}[h]
\begin{center}
\centerline{%
      \includegraphics[width=0.5\textwidth]{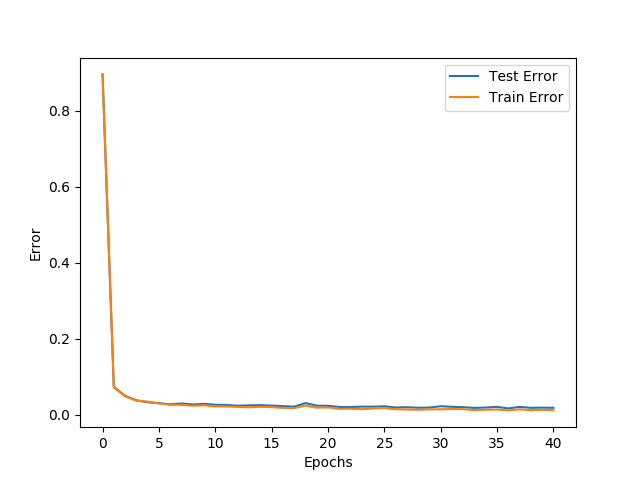}%
    \includegraphics[width=0.5\textwidth]{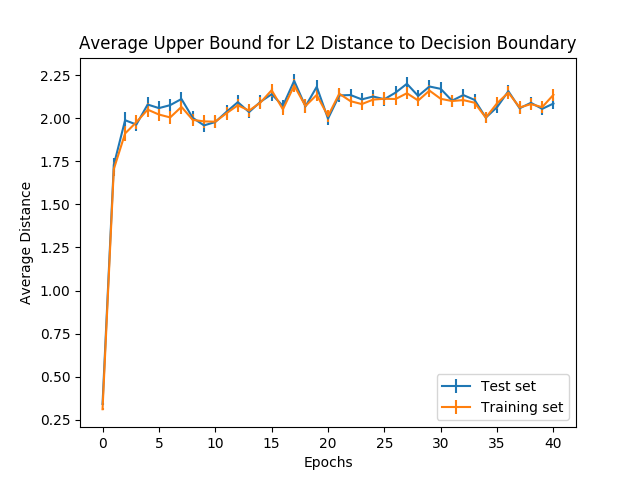}%
      }%
\end{center}
   \caption{(a) The train and test errors also converge quickly to near optimal performance for adversarial training on \textsc{mnist}. \newline
   (b) $d_2^{avg}$ is rather increasing during adversarial training and nearly twice as high as for standard training which is evidence for observation 2.} 
 \label{fig: mnist_results_adversarial_1}
\end{figure*}

\begin{figure*}[h]
\begin{center}
\centerline{%
      \includegraphics[width=0.5\textwidth]{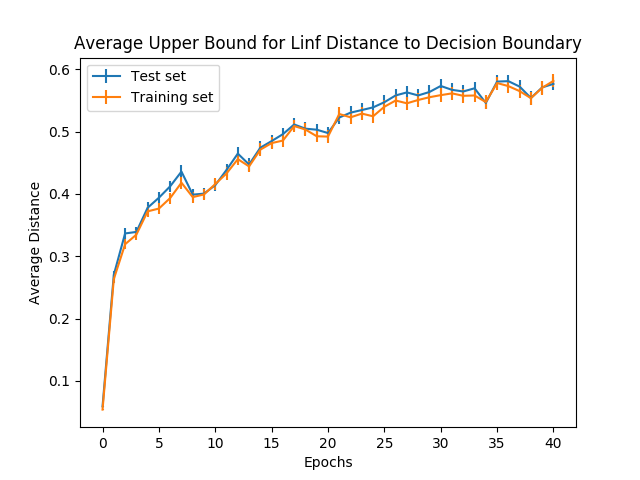}%
      }%
\end{center}
   \caption{$d_{\infty}^{avg}$ is steadily increasing for PGD adversarial training.} 
 \label{fig: mnist_results_adversarial_2}
\end{figure*}

\begin{figure*}[h]
\begin{center}
\centerline{%
      \includegraphics[width=0.5\textwidth]{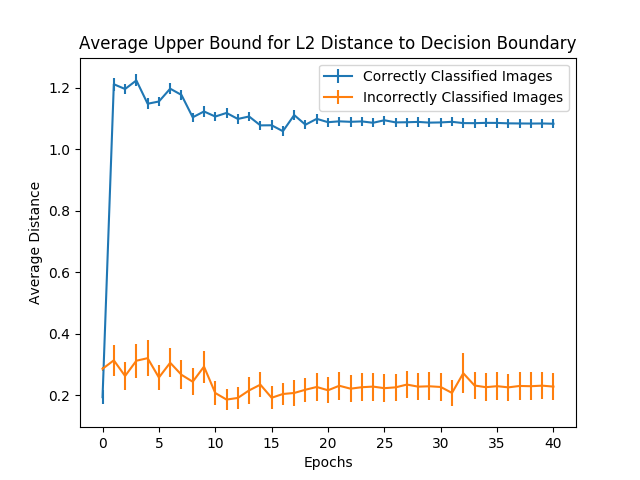}%
    \includegraphics[width=0.5\textwidth]{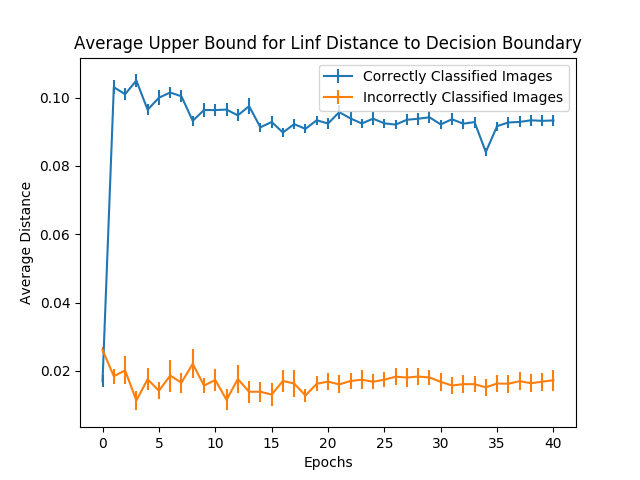}%
      }%
\end{center}
   \caption{(a) and (b): For a dense network on \textsc{Fashion Mnist} with standard training, 
   $d_2^{avg}$ and $d_{\infty}^{avg}$ are much higher for correctly classified data points than for incorrectly classified ones supporting observation 3.} 
 \label{fig: mnist_results_correct_incorrect}
\end{figure*}

For \textsc{mnist} we used a dense network architecture in contrast to the \textsc{cnn} architecture which we used for \textsc{fashion-mnist}.
In Figure~\ref{fig: mnist_results_vanilla} we see that observation 1 is supported even though the network quickly achieves near optimal test error.
Figure~\ref{fig: mnist_results_adversarial_1} supports observation 2 showing that $d_2^{avg}$ is nearly twice as high for adversarial training 
as for standard training. In Figure~\ref{fig: mnist_results_adversarial_2} we even see a steady increase for $d_{\infty}^{avg}$ during 
adversarial training.
Finally, Figure~\ref{fig: mnist_results_correct_incorrect} is evidence for observation 3. Notice that the standard error generally is larger for incorrectly classified 
samples than for correctly classified ones, since there are far fewer of them due to the small error on train and test set.

\begin{figure*}[h]
\begin{center}
\centerline{%
   \includegraphics[width=0.35 \textwidth]{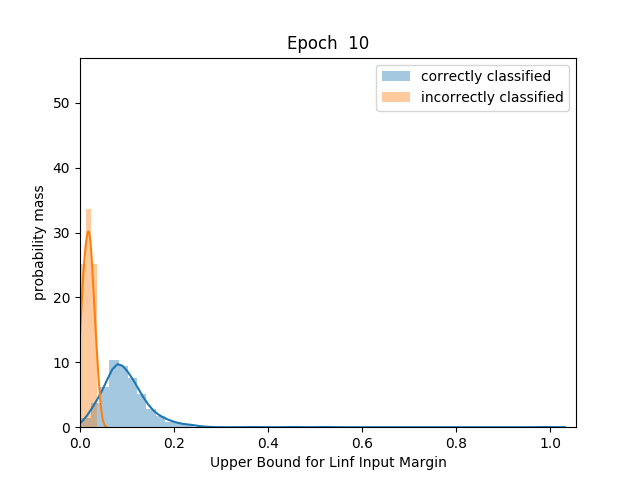}%
    \includegraphics[width=0.35 \textwidth]{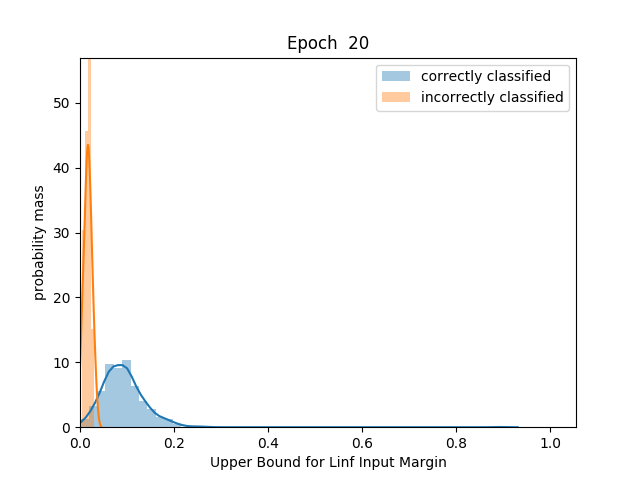}%
    \includegraphics[width=0.35 \textwidth]{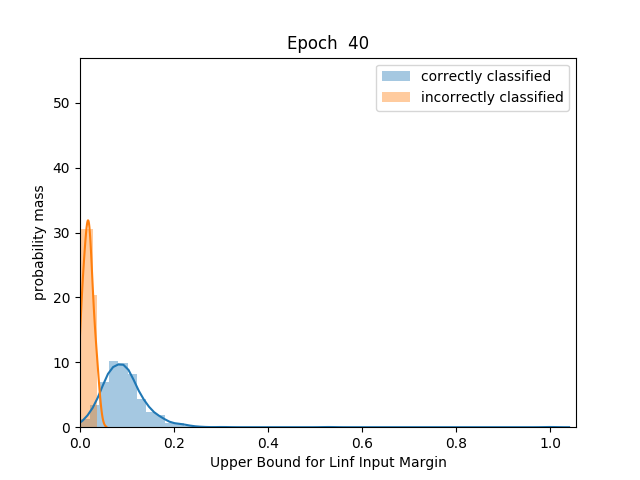}%
    }%
\centerline{%
\includegraphics[width=0.35 \textwidth]{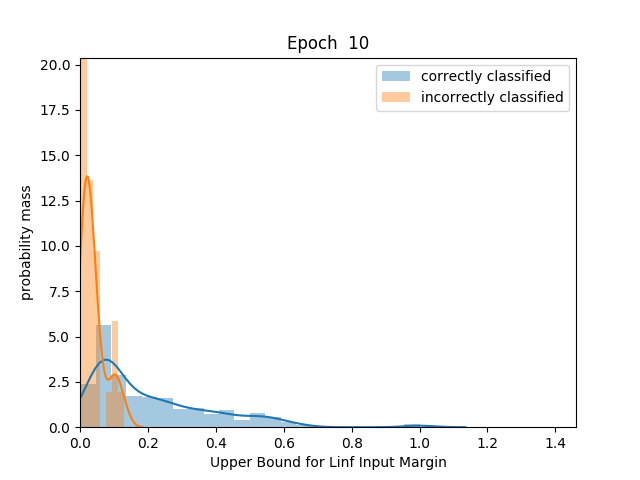}%
\includegraphics[width=0.35 \textwidth]{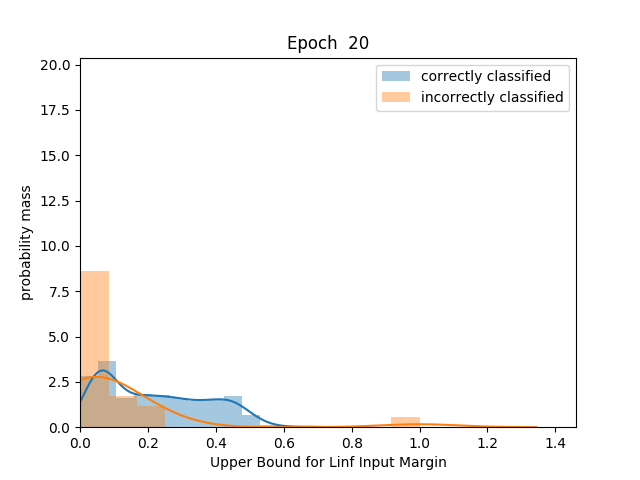}%
\includegraphics[width=0.35 \textwidth]{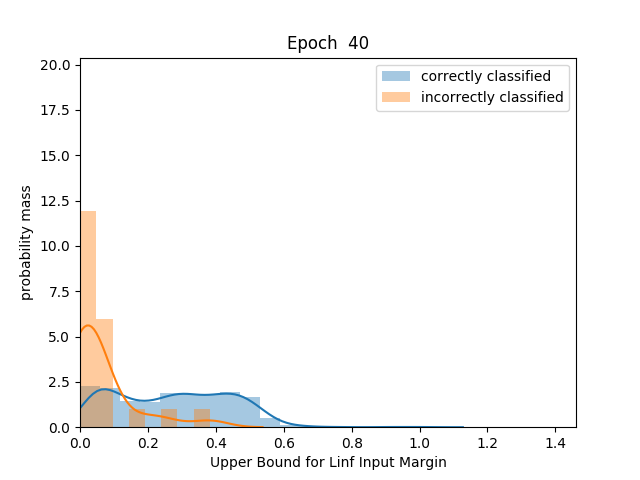}%
}%
\end{center}
   \caption{Sample $d_{\infty}^{avg}$ distributions for $1000$ training images and the 
   training epochs $10, 20$ and $40$: (1) First row: $d_{\infty}^{avg}(x)$ distribution for the 
   classically trained dense network on \textsc{mnist}; (2) Second row: $d_{\infty}(x)^{avg}$ 
   distribution for the adversarially trained dense network on \textsc{mnist}.} 
 \label{fig: mnist_distributions}
\end{figure*}
\clearpage

\section{Work in Progress Notice}
This document is still work in progress and we are planning to release a discussion of the network 
architectures as well the results for \textsc{Cifar-10} in a future version.

\end{document}